\documentclass[final]{cvpr}

\usepackage{times}
\usepackage{epsfig}
\usepackage{graphicx}
\usepackage{amsmath}
\usepackage{amssymb}

\usepackage{caption}
\usepackage{cite}
\usepackage{makecell}
\usepackage{threeparttable}
\usepackage{subfig}
\usepackage{mathrsfs}
\usepackage{booktabs}
\usepackage{multirow}
\usepackage{setspace}
\usepackage{float}
\usepackage{amsmath,amssymb,amsfonts}
\usepackage{algorithmic}
\usepackage{textcomp}

\usepackage[pagebackref=true,breaklinks=true,colorlinks,bookmarks=false]{hyperref}

\def\cvprPaperID{5506} % *** Enter the CVPR Paper ID here
\def\confYear{CVPR 2021}

\begin{document}
	
	%%%%%%%%% TITLE
	\title{U-HRNet: Delving into Improving Semantic Representation of \\ High Resolution Network for Dense Prediction }
	
%	\author{Jian Wang\\
%		Institution1\\
%		Institution1 address\\
%		{\tt\small firstauthor@i1.org}
%		% For a paper whose authors are all at the same institution,
%		% omit the following lines up until the closing ``}''.
%		% Additional authors and addresses can be added with ``\and'',
%		% just like the second author.
%		% To save space, use either the email address or home page, not both
%		\and
%		Second Author\\
%		Institution2\\
%		First line of institution2 address\\
%		{\tt\small secondauthor@i2.org}
%	}
	
	\author{
	Jian Wang\thanks{Equal contribution.} \hspace{0.2cm} Xiang Long$^{\ast}$\hspace{0.2cm} Guowei Chen\hspace{0.2cm} Zewu Wu\hspace{0.2cm} Zeyu Chen\hspace{0.2cm} Errui Ding\\
	Baidu VIS\\
{\tt\small \{wangjian33,chenguowei01,wuzewu,chenzeyu01,dingerrui\}@baidu.com}\\
{\tt\small lxastro0@gmail.com}
    }
	
	\maketitle

	%%%%%%%%% ABSTRACT
	\begin{abstract}
		High resolution and advanced semantic representation are both vital for dense prediction. Empirically, low-resolution feature maps often achieve stronger semantic representation, and high-resolution feature maps generally can better identify local features such as edges, but contains weaker semantic information. 
		Existing state-of-the-art frameworks such as HRNet has kept low-resolution and high-resolution feature maps in parallel, and repeatedly exchange the information across different resolutions. 
		However, we believe that the lowest-resolution feature map often contains the strongest semantic information, and it is necessary to go through more layers to merge with high-resolution feature maps, while for high-resolution feature maps, the computational cost of each convolutional layer is very large, and there is no need to go through so many layers.
		Therefore, we designed a U-shaped High-Resolution Network (U-HRNet), which adds more stages after the feature map with strongest semantic representation and relaxes the constraint in HRNet that all resolutions need to be calculated parallel for a newly added stage. 
		More calculations are allocated to low-resolution feature maps, which significantly improves the overall semantic representation. 
		U-HRNet is a substitute for the HRNet backbone and can achieve significant improvement on multiple semantic segmentation and depth prediction datasets, under the exactly same training and inference setting, with almost no increasing in the amount of calculation.
		Code is available at PaddleSeg\cite{liu2021paddleseg}: \href{https://github.com/PaddlePaddle/PaddleSeg}{https://github.com/PaddlePaddle/PaddleSeg}.
	\end{abstract}
	
	%%%%%%%%% BODY TEXT
	\section{Introduction}
	
	\begin{figure*}[ht]
		\centering  %
		\includegraphics[width=1.0\textwidth]{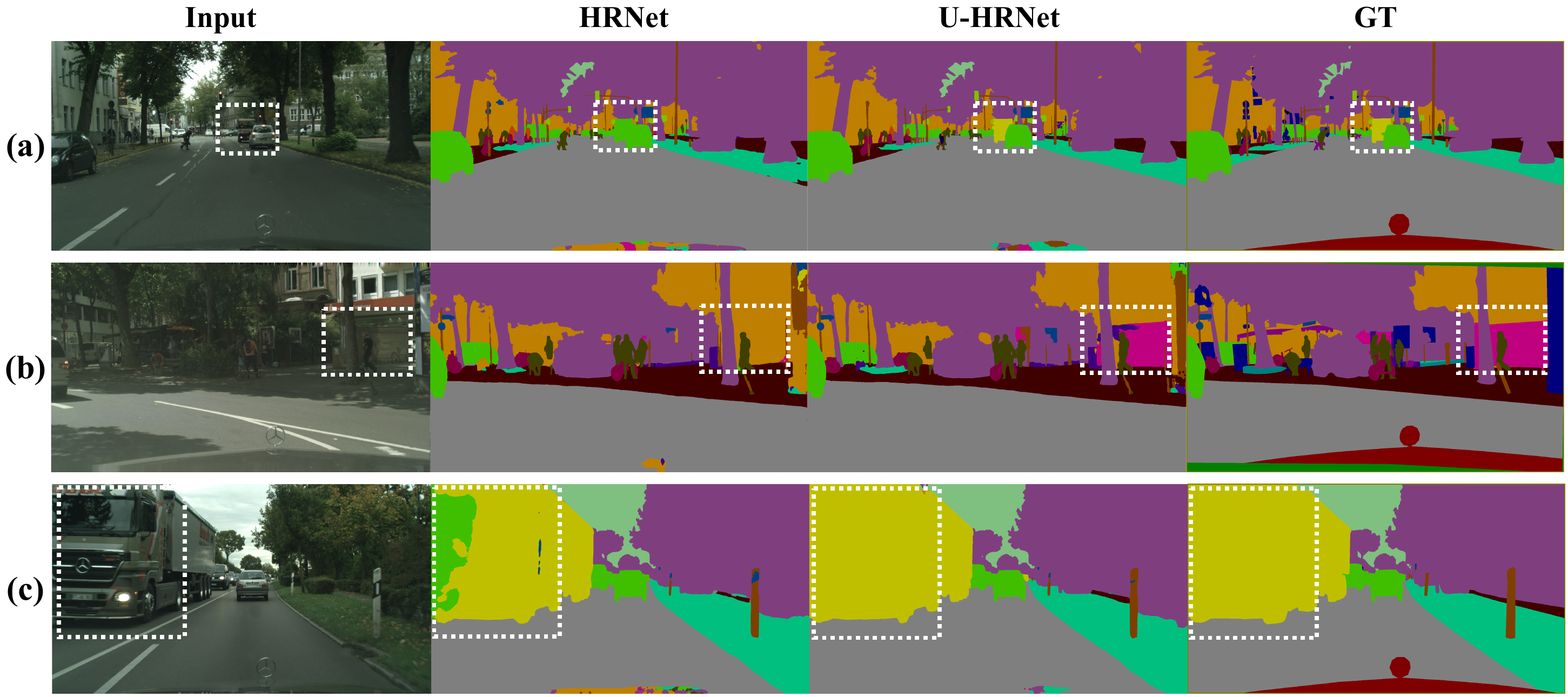}
		\caption{Semantic segmentation results of HRNet and U-HRNet. We use white dashed boxes to mark areas where the results are significantly different.}
		\label{fig.res}
	\end{figure*}
	
	Dense prediction tasks, including semantic segmentation and depth estimation, among many others, are vital part of the visual understanding system. Dense prediction tasks require predicting pixel-level category labels or regressing specific values, which are much more challenging than image-level prediction tasks. 
	Keeping high resolution and strong semantic information at the same time are the key to address dense prediction tasks effectively. High resolution ensures that the final prediction granularity is as close to the pixel level as possible, and more accurate local discrimination can be obtained, such as more accurate edges. Strong semantic information ensures the overall prediction accuracy, especially for instances that are difficult to distinguish or with a relatively large area.
	
	Deep convolutional neural networks, such as U-Net~\cite{unet,zhou2018unetpp}, DeepLab~\cite{ChenPKMY18,deeplabv3}, HRNet~\cite{SunXLW19,hrnet}, which following the design of FCNs \cite{fcn},  have 
	achieved exciting results in dense prediction tasks.  Especially high resolution networks (HRNet)\cite{SunXLW19,hrnet} have achieved state-of-the-art results in dense prediction tasks, such as semantic segmentation, human pose estimation, and so on. HRNet are able to learn high resolution representations, while ensuring the transmission of semantic information between low resolution feature maps and high resolution feature maps at the same time.
	However, we observed that HRNet still has a lot of room for improvement. We can often see that large areas are incorrectly classified. For example, in Figure \ref{fig.res}, the second row shows the results of HRNet. In Figure \ref{fig.res} (a) and (b), the entire instance is misclassified, while in (c), some blocks inside one large-area instance are misclassified. This indicates that semantic representation achieved by HRNet is still not good enough. We suppose that it could be attribute to the macro structure of HRNet, which can be summarized as two aspects below: (i) The final block from the lowest resolution branch of HRNet which has the strongest semantic representation is directly outputted without being fully propagated to higher resolution branches. (ii) The lower resolution branches of HRNet are not deep enough which makes the semantic capacity of network to be limited. While, enlarging the number of modules in the last two stages of HRNet for deeper network is obviously undesirable because of significant increasement in computation cost. U-Net~\cite{unet} alleviates the above two problems to a certain extent. However, in U-Net, only a single resolution is retained in each stage, and there is no fusion between different scales except merging with the residual links. We think that being able to maintain multiple scales in parallel and perform multi-scale fusion at all times is the biggest advantage of HRNet.
	
	With the motivation of improving overall semantic representation of high resolution network without adding extra computation cost, we propose a simple and effective network named as \textbf{U-}shaped \textbf{H}igh \textbf{R}esolution \textbf{Net}work (U-HRNet). It inherits the encoder-decoder structure of U-Net, which is conducive to the embedding propagation from the strongest semantic feature map to the highest resolution feature map. At the same time, it perfectly retains the advantages of HRNet, maintaining multiple scales in parallel and performing multi-scale fusion at all times. Furthermore, it reduces the number of blocks on the high-resolution branches, and reallocates their calculation to the low-resolution ones for larger semantic capacity without adding more computation. As shown in Figure \ref{fig.res}, we can see that our U-HRNet has more advantages than HRNet in the semantic representation of difficult objects and instances with large areas. Fortunately, U-HRNet also works well together with the OCR head proposed recently by~\cite{YuanCW19}, as U-HRNet focuses on improving the semantic capacity of the whole network, which is not overlap with the superiority of OCR which aims to better labeling with the help of semantic relationship between objects and categories.
	
	Thus, the contributions of this paper are following two points. 1) A simple and effective network U-HRNet is proposed, which outperforms HRNet on dense prediction tasks with almost no increasement of computation. 2) U-HRNet togethor with OCR sets new state-of-the-arts on several semantic segmentation datasets.

	\section{Related work}
	
	Semantic segmentation is a pixel-level classification task, but it also requires an in-depth understanding of the semantics of the entire scene in order to assign the correct label to each pixel. Since Long \etal. \cite{fcn} proposed fully convolutional network (FCN), which includes only convolutional layers and can take an image of arbitrary size as inputs, various deep convolutional neural networks (DCNNs)  have been dominate the task of semantic segmentation. 
	To integrate more scene-level semantic context, several approaches \cite{ChenPKMY18,schwing2015fully,lin2016efficient} incorporate probabilistic graphical models into DCNNs. For example,  Chen \etal. \cite{ChenPKMY18} proposed to combine DCNNs with fully connected conditional random fields.  
	Several other works, such as  RefineNet \cite{lin2017refinenet}, ExFuse \cite{zhang2018exfuse} and CCL \cite{ding2018context}, have been proposed to capture rich contextual information from the perspective of multi-scale aggregation.
	
	Instead of using the traditional DCNN backbones\cite{alexnet,googlenet,VGG,ResNet,Densenet} commonly used in other computer vision tasks, developing a backbone network that is more suitable for segmentation can achieve better semantic segmentation performance.
	One popular family of DCNNs for image segmentation is based on the convolutional encoder-decoder architecture \cite{minaee2020image}. 
	U-Net \cite{unet}, and V-Net \cite{vnet}, are two well-known such architectures. 
	They both need to go through multiple downsampling convolutional blocks and then use multiple upsampling deconvolutional blocks to restore the original resolution.
	PSPNet \cite{PSPNet}  and Deeplabv3 series \cite{deeplabv3,deeplabv3plus}  retain multiple spatial resolutions through different sizes of pooling kernels or dilated convolutions with different dilated rates.
	HRNet\cite{SunXLW19,hrnet } is first proposed on the human pose estimation task. It is also very suitable for semantic segmentation tasks and can achieve state-of-the-art results. HRNet direclty maintains high-to-low resolution convolution streams in parallel and repeatedly exchanging the information across resolutions.  
	Our method U-HRNet is build based on HRNet. It inherits the advantages of HRNet, keeping several resolution streams and repeating cross-resolution information transmission. At the same time, we allocate the unnecessary amount of computation on the high-resolution feature map to more meaningful parts to improve the overall semantic presentation. 
	Our method has a certain similarity with U-Net, since they are both U-shaped networks. But different from U-Net, which only maintains a single resolution at each stage,  each stage of U-HRNet will inherits one resolution stream from previous stage, which is more helpful for the fusion between different resolutions.
	
	In addition, there are many methods based on self-attention \cite{vaswani2017attention} to improve the semantic representation. For example, OCNet \cite{YuanCW19} aggregates objects context by applying a self-attention module to compute the similarities of each pixel and all the other pixels. DANet \cite{DANet} and Relational Context-aware Network \cite{Lichao2019A} explore contextual dependencies in both spatial and channel dimensions.  HCNet\cite{hcnet} proposes a method to captures the dependencies between pixels within each homogeneous region and continues to model the correlation between different regions.  
	Since our work is mainly focus on improving the backbone structure, we did not add these modules to our model for pure comparison.
	In fact, most of these backbone-independent improvements can be directly applied to U-HRNet to achieve better results.
	
	Depth estimation is a pixel-level regression task, which also requires global understanding of the whole scene. Early works~\cite{saxena2007learning} depend on hand-crafted features to address the problem. The performance has been boosted significantly since using deep learning to predict depth maps \cite{eigen,CVPR15Liu,eigen2015predicting,liu2015learning,wang2015towards,chakrabarti2016depth,laina2016deeper,li2017two,alhashim2018high,li2018deep,fu2018deep,fei2019geo,wei2019enforcing,hu2019revisiting}. As in semantic segmentation, various backbones \cite{unet,deeplabv3,hrnet} have also been used in depth estimation. In this paper, we also show that U-HRNet can achieve better results in depth estimation task.
	
	\vspace{-2mm}
	\section{U-Shape High-Resolution Network}
	
	\begin{figure*}[t]
		\footnotesize
		\centering
		\subfloat[]{\includegraphics[width=0.49\textwidth]{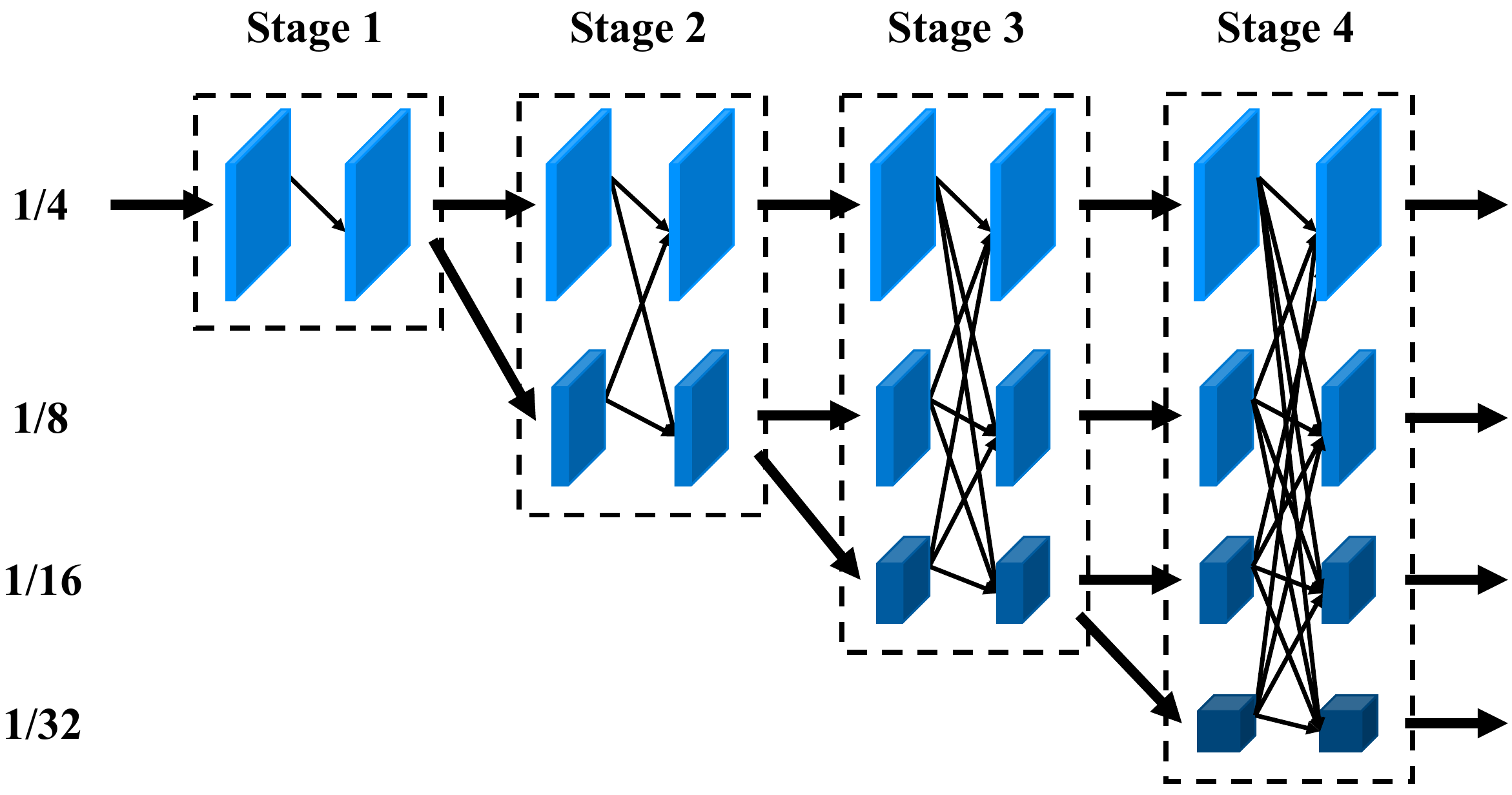}} ~~~~
		\subfloat[]{\includegraphics[width=0.49\textwidth]{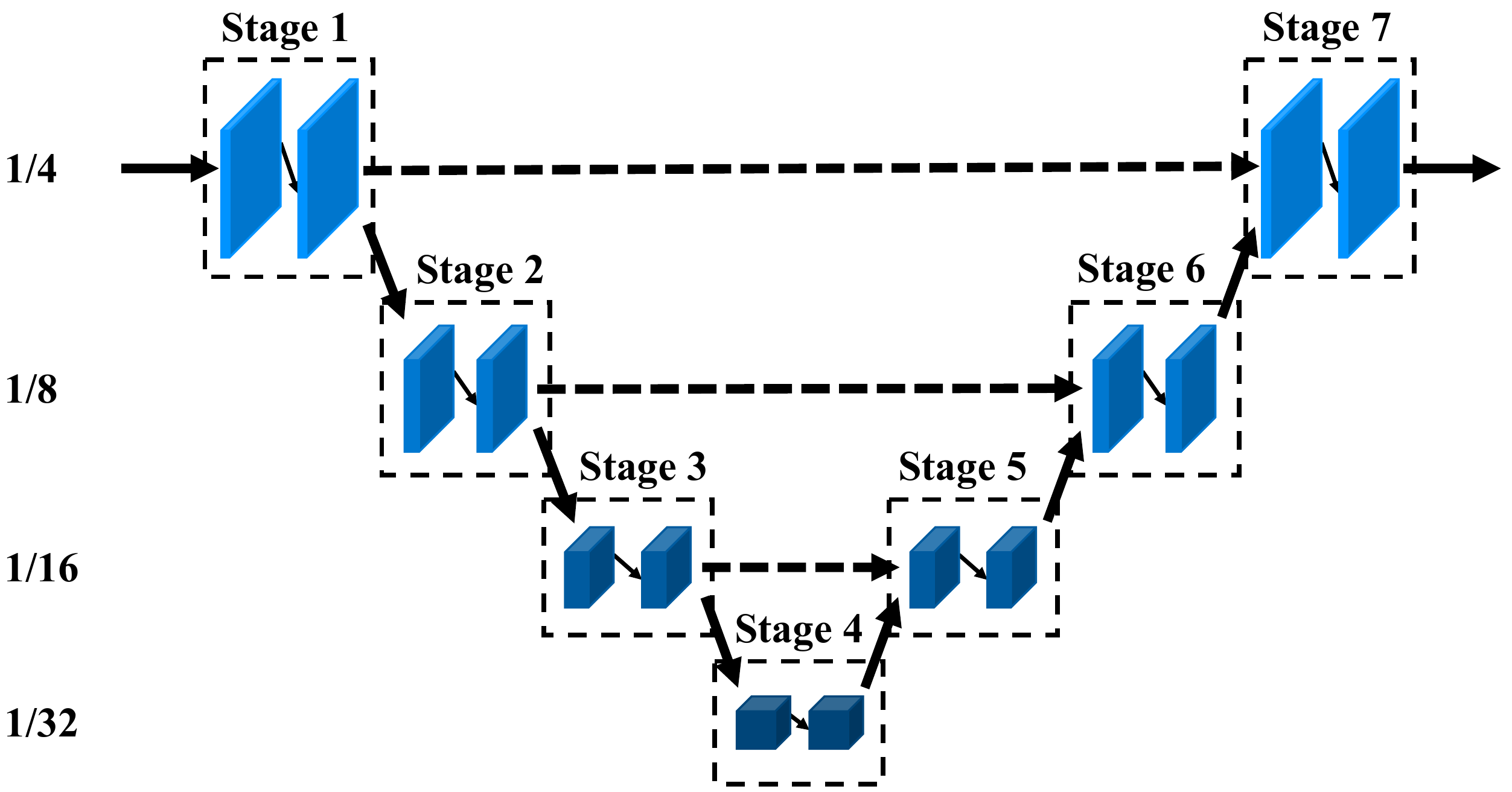}} \\
		\subfloat[]{\includegraphics[width=0.85\textwidth]{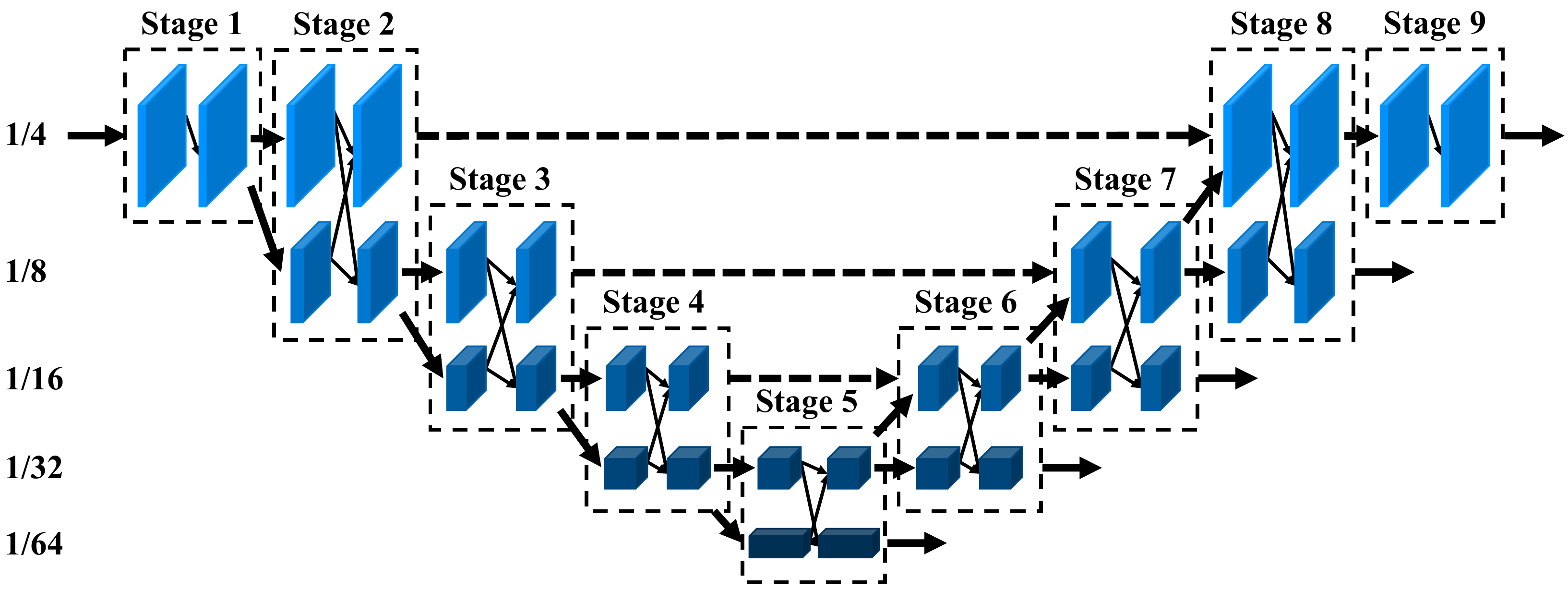}}
		\caption{
			Illustrating the network architecture of different methods. 
			(a) HRNet consists of parallel high-to-low resolution sub-networks with repeated hr-modules to exchange information across multi-resolution feature maps.
			(b) U-Net only maintains a single resolution at each stage.
			(c) U-HRNet is a U-shape network in macro scope. Each stage is composed of several hr-modules formed by no more than two resolution branches.
		}
		\label{fig:architectures}
		\vspace{-3mm}
	\end{figure*}
	
	\subsection{Review of HRNet} \label{subsection:review}
	HRNet is an excellent neural network first proposed in ~\cite{SunXLW19} for human pose estimation. After that, ~\cite{hrnet} further proved that HRNet could work very well on many other tasks, such as object detection, semantic segmentation. Thus it can be seen that, HRNet is strong not only at high-level semantic representation but also low-level spatial detail. As shown in Figure~\ref{fig:architectures} (a), the 1/4 resolution is consistent from the beginning to the end of network, and in pace with the increasing of network of depth, more lower resolutions are added for semantic representation learning, thus improving the high-resolution representation with multi-resolution fusion.
	
	However, HRNet may not be perfect for some dense prediction tasks. For instance, semantic segmentation is a typical dense classification task, it is important to introduce high-level global information for helping a pixel to predict its semantic category. From this point of view, we find HRNet has several drawbacks below: (i) The last block of 1/32 resolution branch which has the strongest semantic representation is directly outputted without being fully utilized. (ii) The calculation allocation among high and low resolution branches is not optimized, and the low resolution branches with strong semantic representation should be paid more attention.
	
	\subsection{Architecture of U-HRNet}
	In order to alleviate the shortcomings mentioned above, we propose a simple and effective high resolution network named U-HRNet, which improves the semantic representation of high resolution output by restructuring the macro layout of HRNet. We will describe the details of U-HRNet from the following three perspectives: main body, fusion module and representation head.
	
	\vspace{-2mm}
	\subsubsection{Main Body}
	Following HRNet, we input the image into a stem block decreasing the resolution to 1/4, and the main body outputs the feature map with same resolution as 1/4. Figure~\ref{fig:architectures} (c) illustrates the main body of U-HRNet. Alike with U-Net, whose layout indicated in Figure~\ref{fig:architectures} (b), the main body appears to be a U-Shape network in macro scope, while in micro scope, it is composed of several hr-modules proposed in ~\cite{SunXLW19,hrnet}. However, each hr-module is formed by no more than two resolution branches. This designing manner is targeted at dealing with the drawbacks of HRNet mentioned in Section~\ref{subsection:review}. The details of restructuring are elaborated as below. Firstly, we remove high-resolution branches of the last two stages of HRNet (1/4 resolution branches of stage3 and stage4, 1/8 resolution branch of stage4), which make a lot of calculations be released. Then, for improving semantic representation of the high-resolution output, we add several stages after the lowest resolution stage. These stages gradually upsample the feature maps and merge with the features from previous stages. That makes the feature with the strongest semantic representation outputted by the lowest resolution stage can be merged with low-level high-resolution features more earlier, thus the successive stages are able to infer the spatial details more precise by analyzing the strongest representation sufficiently. Finally, we rearrange the hr-modules in different stages. We increase modules in low-resolution stages while reduce in high-resolution stages, which improving the semantic representation to a higher degree. In addition, we add a stage with 1/32 and 1/64 resolution branches for producing more informative semantic representation without need of adding extra higher resolution branches. Analogous to U-Net, several shortcuts are placed span the depth direction of network, which connect the stage2 and stage8, stage3 and stage7, stage4 and stage6 respectively. These shortcuts make the network can take advantages of both high-level features and low-level features, and meanwhile, enable the gradient to propagate directly to the previous stages. 
	
	\subsubsection{Fusion Module} \label{subsubsection:fusion}
\begin{figure}[t]
		\centering  %
		\includegraphics[width=0.45\textwidth]{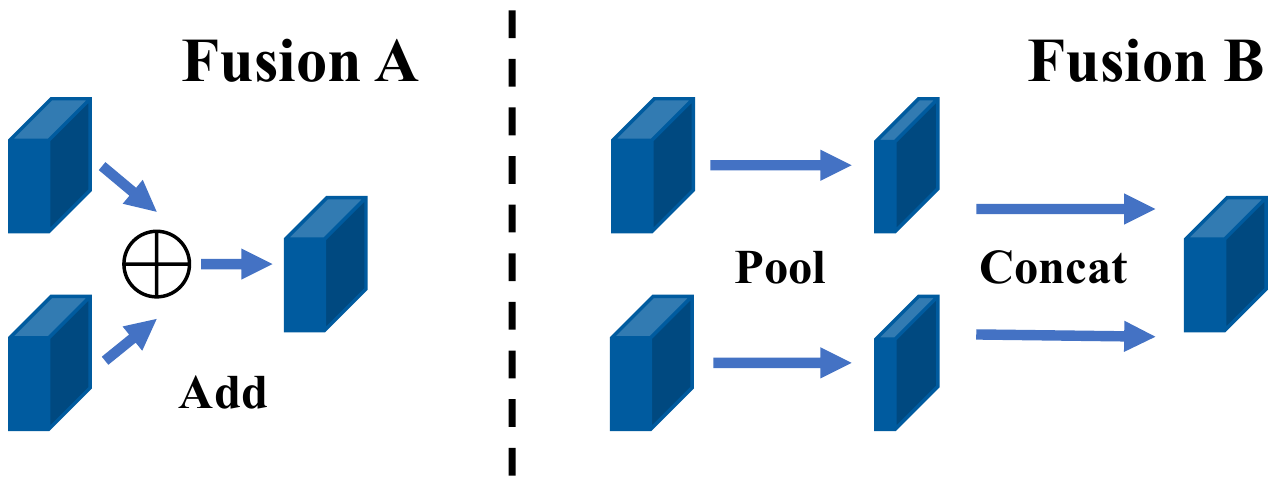}
		\caption{Fusion modules: Fusion A is the fusion module in HRNet\cite{hrnet}, which directly adds the two input features. The proposed Fusion B first pools the two input features at channel dimension with a kernel size of 2 and then concatenates them together. }
		\label{fig:components}
	\end{figure}

	Corresponding to the shortcuts in main body, there are three fusion modules before stage8, stage7 and stage6, which merge the low-level features outputted by the higher resolution branches of stage2, stage3 and stage4 with the upsampled features from the higher resolution branches of stage7, stage6 and stage5 respectively. Intuitively, we can apply the method of fusion used in high resolution module for simplicity, which adding the two input features and then conducting a ReLU function for activation, as exhibited in Fusion A from Figure~\ref{fig:components}. However, we suppose that concatenating two input features can enhance the connectivity of network according to the fusion method of U-Net. Therefore, we first pool the two input features at channel dimension with a kernel size of 2 and then concatenate them among channel as the output feature, as exhibited in Fusion B from Figure~\ref{fig:components}.
	
	\subsubsection{Representation Head}
	For the representation head, we follow HRNetV2 basicly. The outputted multi-resolution features are from the lower resolution branches of stage5, stage6, stage7, stage8 and stage9. However, due to the added 1/64 resolution, the numbers of input channel of the convolution in the U-HRNet representation head is double to the HRNetV2. For sake of keeping a similar computation cost to the representation head of HRNetV2, we pass the multi-resolution features through a pooling operation with kernel size as 2, and then concatenate them together among the channel dimension as the input of representation head.
	
	\subsection{Instantiation}
	The main body of U-HRNet contains $9$ stages with $5$ resolution streams. The resolutions are $1/4$, $1/8$, $1/16$, $1/32$, $1/64$. The first stage contains $1$ one-branch hr-module formed by $4$ bottleneck residual blocks where each block has a width of $64$, and is followed by one $3\times3$ convolution changing the width of feature map to $C$, which is denoted as the width of $1/4$ resolution stream. The $2$nd to $8$th stages contain $1$, $5$, $2$, $2$, $1$, $1$, $1$ hr-modules, respectively. And all of these modules are composed by two branches, each branch in these modules consist of $4$ basic residual blocks. Like the first stage, the last stage contains $1$ one-branch hr-module too, while this module is formed by $4$ basic residual blocks. Finally, the widths of the convolutions of the five resolution streams are $C$, $2C$, $4C$, $8C$ and $16C$ respectively. The layout of U-HRNet is depicted in Figure~\ref{fig:architectures} (c). Besides, for U-HRNet-small, there are two points of difference comparing to U-HRNet, which are (i) the third stage is consist of $2$ hr-modules while the other stages stay same with U-HRNet, (ii) all the branches in the hr-modules from U-HRNet-small contain $2$ bottleneck blocks or basic blocks.
	
	\subsection{Analysis}
	Beside the first and last stage, in U-HRNet, two-branches hr-module is mainly adopted as the basic unit for composing the network instead of the multi-resolution (two or more) parallel convolution used in HRNet and one-branch convolution sequence in U-Net. It brings several benefits which helping to improve the semantic representation. Compare to HRNet, the two-branches hr-module relaxes the limitation that all resolutions need to be calculated parallel for a certain stage while without lost of the advantage of multi-resolution inference. This allows U-HRNet can attach more calculation on low resolution branches than HRNet and further improve the strongest semantic representation. Compare to U-Net, the two-branches hr-module is superior to one-branch convolution sequence on multi-scale representation learning significantly. Moreover, when stepping into next resolution, either downsampling or upsampling, U-HRNet still keeps one of the previous resolutions ongoing and fuses the features of two resolutions continuously. That makes the network can fully take advantage of the information learned before while avoiding the lost of spatial or semantic knowledge caused by changing resolution.
	
	\section{Semantic Segmentation}
	Semantic segmentation is a problem of predicting class label for each pixel in a image. It is a basic and important dense task in scene understanding. Here, we conduct extensive experiments on three popular datasets of semantic segmentation, including Cityscapes, ADE20K and LIP, and then report state-of-the-art results. In addition, we also take experiment on a medical benchmark, Synapse multi-organ CT, and then show competitive results.
	
	\subsection{Datasets and Evaluation Metrics}
	
	\noindent\textbf{Cityscapes.} 
	The Cityscapes dataset~\cite{cityscapes} is a large-scale dataset used for semantic urban scene understanding. It contains $5,000$ images with fine annotations and $20,000$ images with coarse annotations, collected from $50$ different cities. This dataset includes a total of $30$ classes, $19$ of which are used for actual training and validation. It is noted that in our experiments, we only use $5,000$ images with fine annotations as our dataset, which is divided into $2,975$, $500$ and $1,525$ images for training, validation and online testing. The mean IoU of all classes (mIoU) is used as the major score on this dataset, and in addition, three extra scores are reported on \textit{test} set: IoU category(cat.), iIoU class(cla.) and iIoU category(cat.).
	
	\noindent\textbf{ADE20K.}
	The ADE20k dataset~\cite{zhou2017scene} is a challenging scene parsing dataset which has 150 classes with 1038 image-level labels, and varies in different scenes and scales. It is splitted as training/validation/testing sets with 20K/2K/3K images included respectively. The mIoU is applied as main score on this dataset.
	
	\noindent\textbf{LIP.} 
	The LIP dataset~\cite{GongLSL17} contains $50,462$ elaborately annotated human images, in which $30,462$ images are used for training, and $10,000$ images are used for testing. This dataset contains $20$ categories ($19$ human part labels and $1$ background label). The mIoU is the main metric for this dataset, and pixel accuracy(acc.) and average(avg.) accuracy(acc.) are also reported.
	
	\noindent\textbf{Synapse multi-organ CT.} Here, we adopt the same dataset setting following~\cite{chen2021transunet}, where $30$ abdominal CT scans are used, with totally $3779$ axial contrast-enhanced abdominal clinical CT images. There are $85 \sim 198$ slices of $512 \times 512$ pixels in each CT volume which has spatial resolution of $([0.54 \sim 0.54] \times [0.98 \sim 0.98] \times [2.5 \sim 5.0]) mm^3$ for a voxel. We also apply the average DSC and average Hausdorff Distance (HD) on 8 abdominal organs as evaluation metrics as same as~\cite{chen2021transunet}.

	\subsection{Implementation Details}
	For the training and testing settings, we follow~\cite{hrnet, YuanCW19} upon the experiment of U-HRNet and U-HRNet+OCR on Cityscapes, ADE20K and LIP. While, on Synapse multi-organ CT, we follow~\cite{chen2021transunet}.
	
	\subsection{Experimental Results}
	
	\subsubsection{Ablation Study}
	
	We use HRNetV$2$-W$18$-small-v2 as the baseline structure and perform ablation study for several different structure configurations of U-HRNet. And all of the experiments are training from scratch. For convenience, we encode the network structure with a sequence of numbers. Each number in the sequence represents a stage, and the number itself is equal to the amount of hr-modules in this stage. For instance, "$1$\begin{tiny}$\searrow$\end{tiny}$1$\begin{tiny}$\searrow$\end{tiny}$2$\begin{tiny}$\searrow$\end{tiny}$5$\begin{tiny}$\nearrow$\end{tiny}$1$\begin{tiny}$\nearrow$\end{tiny}$1$\begin{tiny}$\nearrow$\end{tiny}$1$" means a network which has $7$ stages and the amount of hr-modules in each stage is $1$, $1$, $2$, $5$, $2$, $1$, $1$ respectively. In addition, in that sequence, $\searrow$ means conducting downsampling between two neighbouring stages, while upsampling for $\nearrow$. Specially, without extra claim, the first stage and last stage of the sequence is consist of one-branch hr-modules, while two-branches modules for the other stages. Otherwise, if a sequence is ended with $^=$, the last stage of this sequence is consist of two-branches hr-modules. More details are shown in supplementary material.
	
	Table~\ref{tab:ablation} illustrates the computation cost(GFLOPs) and performance(mIoU) of HRNetV$2$-W$18$-small-v2, U-HRNet-W$18$-small and its several variations. From the comparison between U-HRNet-W$18$-small-va and HRNetV$2$-W$18$-small-v2, we find that the high resolution branches of the last two stages in HRNet are not very necessary. While making the small-scale branches deeper and placing several upsampling hr-modules at the back of network as a decoder both improve the performance greatly, as shown from U-HRNet-W$18$-small-va to U-HRNet-W$18$-small-ve. And so on, we replace two of the hr-modules which lower resolutions are 1/32 with two modules which lower resolutions are 1/64 base on U-HRNet-W$18$-small-ve to boost the performance of network even further. Finally, we achieve 75.1 mIoU on Cityscapes val which outperforms the HRNetV$2$-W$18$-small-v2 baseline about 5 percent. In addition, different distribution of hr-modules among all the stages and different fusion modules are studied. U-HRNet-W$18$-small-vf and U-HRNet-W$18$-small-vg are two variant structures which previous and later stages are enlarged respectively. U-HRNet-W$18$-small-vh use fusion A instead of fusion B as the fusion module. The results in Table~\ref{tab:ablation} exhibit that all these variations are not as competitive as U-HRNet-W$18$-small.
	
	\begin{table}[t]
		\centering
		\caption{Ablation study on Cityscapes \texttt{val}. The GFLOPs is calculated on the input size $1024 \times 2048$.
		}
		\label{tab:ablation}
		\vspace{-.2cm}
		\resizebox{0.48\textwidth}{!}{
			\begin{tabular}{l|cc|c}
				\hline & structure & GFLOPs & mIoU \\
				\hline
				\hline
				HRNetV$2$-W$18$-small-v2 & - & $71.6$ & $70.1$ \\
				\hline
				U-HRNet-W$18$-small-va & $1$\begin{tiny}$\searrow$\end{tiny}$1$\begin{tiny}$\searrow$\end{tiny}$3$\begin{tiny}$\searrow$\end{tiny}$2^=$ & $58.6$ & $71.4$ \\
				U-HRNet-W$18$-small-vb & $1$\begin{tiny}$\searrow$\end{tiny}$1$\begin{tiny}$\searrow$\end{tiny}$3$\begin{tiny}$\searrow$\end{tiny}$5^=$ & $67.7$ & $72.8$ \\
				U-HRNet-W$18$-small-vc & $1$\begin{tiny}$\searrow$\end{tiny}$1$\begin{tiny}$\searrow$\end{tiny}$3$\begin{tiny}$\searrow$\end{tiny}$7^=$ & $73.8$ & $73.3$ \\
				U-HRNet-W$18$-small-vd & $1$\begin{tiny}$\searrow$\end{tiny}$1$\begin{tiny}$\searrow$\end{tiny}$2$\begin{tiny}$\searrow$\end{tiny}$5$\begin{tiny}$\nearrow$\end{tiny}$1^=$ & $67.7$ & $73.5$ \\
				U-HRNet-W$18$-small-ve & $1$\begin{tiny}$\searrow$\end{tiny}$1$\begin{tiny}$\searrow$\end{tiny}$2$\begin{tiny}$\searrow$\end{tiny}$5$\begin{tiny}$\nearrow$\end{tiny}$1$\begin{tiny}$\nearrow$\end{tiny}$1$\begin{tiny}$\nearrow$\end{tiny}$1$ & $72.2$ & $74.7$ \\
				\hline
				U-HRNet-W$18$-small-vf & $1$\begin{tiny}$\searrow$\end{tiny}$1$\begin{tiny}$\searrow$\end{tiny}$4$\begin{tiny}$\searrow$\end{tiny}$1$\begin{tiny}$\searrow$\end{tiny}$1$\begin{tiny}$\nearrow$\end{tiny}$1$\begin{tiny}$\nearrow$\end{tiny}$1$\begin{tiny}$\nearrow$\end{tiny}$1$\begin{tiny}$\nearrow$\end{tiny}$1$ & $73.1$ & $73.0$ \\
				U-HRNet-W$18$-small-vg & $1$\begin{tiny}$\searrow$\end{tiny}$1$\begin{tiny}$\searrow$\end{tiny}$2$\begin{tiny}$\searrow$\end{tiny}$1$\begin{tiny}$\searrow$\end{tiny}$1$\begin{tiny}$\nearrow$\end{tiny}$1$\begin{tiny}$\nearrow$\end{tiny}$2$\begin{tiny}$\nearrow$\end{tiny}$2$\begin{tiny}$\nearrow$\end{tiny}$1$ & $73.1$ & $73.3$ \\
				\hline
				U-HRNet-W$18$-small-vh & $1$\begin{tiny}$\searrow$\end{tiny}$1$\begin{tiny}$\searrow$\end{tiny}$2$\begin{tiny}$\searrow$\end{tiny}$2$\begin{tiny}$\searrow$\end{tiny}$2$\begin{tiny}$\nearrow$\end{tiny}$1$\begin{tiny}$\nearrow$\end{tiny}$1$\begin{tiny}$\nearrow$\end{tiny}$1$\begin{tiny}$\nearrow$\end{tiny}$1$ & $73.1$ & $73.9$ \\
				\hline
				U-HRNet-W$18$-small & $1$\begin{tiny}$\searrow$\end{tiny}$1$\begin{tiny}$\searrow$\end{tiny}$2$\begin{tiny}$\searrow$\end{tiny}$2$\begin{tiny}$\searrow$\end{tiny}$2$\begin{tiny}$\nearrow$\end{tiny}$1$\begin{tiny}$\nearrow$\end{tiny}$1$\begin{tiny}$\nearrow$\end{tiny}$1$\begin{tiny}$\nearrow$\end{tiny}$1$ & $73.1$ & $\mathbf{75.1}$ \\
				\hline
			\end{tabular}
		}
		\vspace{-.2cm}
	\end{table}

	\begin{table}[t]
		\centering
		\caption{Semantic segmentation results on Cityscapes \texttt{val} (single-scale and no flipping). The GFLOPs is calculated on the input size $1024 \times 2048$. The U-HRNet-W$48$ performs better than HRNet with almost same GFLOPs and other representative contextual methods (Deeplab and PSPNet) with much smaller GFLOPs. (D- = Dilated-)
		}
		\label{tab:cityscapevalresults}
		\vspace{-.2cm}
		\resizebox{0.48\textwidth}{!}{
			\begin{tabular}{l|lc|c}
				\hline%
				& backbone & GFLOPs & mIoU\\
				\hline
				\hline
				MD(Enhanced)~\cite{xie2018improving} & MobileNetV$1$ & $240.2$ & $67.3$ \\
				ResNet18(1.0) & ResNet18 & 477.6 & 69.1 \\
				MobileNetV$2$Plus~\cite{liu2018lightnet} & MobileNetV$2$ & $320.9$ & $70.1$ \\
				HRNetV$2$~\cite{hrnet} & HRNetV$2$-W$18$-small-v$1$ & $31.1$ & $70.3$ \\
				\hline
				HRNetV$2$~\cite{hrnet} & HRNetV$2$-W$18$-small-v$2$ & $71.6$ & $76.3$ \\
				U-HRNet & U-HRNet-W$18$-small & $73.1$ & $\mathbf{78.5}$ \\
				\hline
				\hline
				U-Net++~\cite{zhou2018unetpp} & ResNet-$101$  & $748.5$ & $75.5$ \\
				D-ResNet~\cite{ResNet} & D-ResNet-$101$  & $1661.6$ & $75.7$ \\
				DeepLabv3~\cite{deeplabv3} & D-ResNet-$101$  & $1778.7$ & $78.5$ \\
				DeepLabv3+~\cite{deeplabv3plus} & D-Xception-$71$ & $1444.6$ & $79.6$ \\
				PSPNet~\cite{PSPNet} & D-ResNet-$101$ & $2017.6$ & $79.7$ \\
				\hline
				HRNetV$2$~\cite{hrnet} & HRNetV$2$-W$48$ & $696.2$ & $81.1$ \\
				U-HRNet & U-HRNet-W$48$ & $698.6$ & $\mathbf{81.9}$ \\
				\hline
				HRNetV$2$+OCR~\cite{YuanCW19} & HRNetV$2$-W$48$ & $1206.3$ & $81.6$ \\
				U-HRNet+OCR & U-HRNet-W$48$ & $1222.3$ & $\mathbf{82.3}$ \\
				\hline
			\end{tabular}
		}
		\vspace{-.2cm}
	\end{table}
	
	\begin{table}[t]
		\centering 
		\caption{Semantic segmentation results on Cityscapes \texttt{test} (multi-scale and flipping). We use U-HRNet-W48, whose computation complexity is comparable to HRNetV$2$-W$48$ and dilated-ResNet-$101$ (D- = Dilated-) based networks, for comparison. Our results are superior in terms of the four evaluation metrics.
		}
		\label{tab:cityscaperesults}
		\vspace{-.2cm}
		\resizebox{0.48\textwidth}{!}{
			\begin{tabular}{l|l|cccc}
				\hline%
				& backbone & mIoU  & iIoU cla. & IoU cat. & iIoU cat.\\
				\hline
				\hline
				\multicolumn{3}{l}{\emph {Model learned on the \texttt{train} set}}\\
				\hline
				PSPNet~\cite{PSPNet} & D-ResNet-$101$ & $78.4$ & $56.7$ & $90.6$  & $78.6$ \\
				PSANet~\cite{ZhaoZLSLLJ18} & D-ResNet-$101$ & $78.6$ & - & - & - \\
				PAN~\cite{LiXAW18} & D-ResNet-$101$ & $78.6$ & - & - & - \\
				AAF~\cite{KeHLY18} & D-ResNet-$101$ & $79.1$ & - & - & -\\
				\hline
				HRNetV$2$~\cite{hrnet} & HRNetV$2$-W$48$ & $80.4$ & $59.2$ & $91.5$ & $80.8$\\
				U-HRNetV$2$ & U-HRNetV$2$-W$48$ & $\mathbf{81.6}$ & $\mathbf{60.8}$ & $\mathbf{92.0}$ &
				$\mathbf{81.2}$\\
				\hline
				\hline
				\multicolumn{3}{l}{
					\emph {Model learned on the \texttt{train+val} set}}\\
				\hline
				GridNet~\cite{FourureEFMT017} & - & $69.5$ & $44.1$ & $87.9$ & $71.1$\\
				LRR-4x~\cite{GhiasiF16} & - & $69.7$ & $48.0$ & $88.2$ & $74.7$\\
				DeepLab~\cite{ChenPKMY18} & D-ResNet-$101$ & $70.4$ & $42.6$ & $86.4$ & $67.7$\\
				LC~\cite{LiLLLT17}& - & $71.1$ & - & - & - \\
				Piecewise~\cite{lin2016efficient}& VGG-$16$ & $71.6$ & $51.7$ & $87.3$ & $74.1$\\
				FRRN~\cite{PohlenHML17}& - & $71.8$ & $45.5$ & $88.9$ & $75.1$\\
				RefineNet~\cite{lin2017refinenet}& ResNet-$101$ & $73.6$ & $47.2$ & $87.9$ & $70.6$\\
				PEARL~\cite{JinLXSLYCDLJFY17} & D-ResNet-$101$ & $75.4$ & $51.6$ & $89.2$ & $75.1$ \\
				DSSPN~\cite{LiangZX18} & D-ResNet-$101$ & $76.6$ & $56.2$ & $89.6$ & $77.8$\\
				LKM~\cite{PengZYLS17}& ResNet-$152$ & $76.9$ & - & - & - \\
				DUC-HDC~\cite{WangCYLHHC18}& - & $77.6$ & $53.6$ & $90.1$ & $75.2$\\
				SAC~\cite{ZhangTZLY17} & D-ResNet-$101$ & $78.1$ & - & - & - \\
				DepthSeg~\cite{KongF18} & D-ResNet-$101$ & $78.2$& - & - & - \\
				ResNet38~\cite{WuSH16e} & WResNet-38 &$78.4$ &$59.1$ &$90.9$ &$78.1$ \\
				BiSeNet~\cite{YuWPGYS182} & ResNet-$101$ & $78.9$ & - & - & - \\
				DFN~\cite{YuWPGYS18} & ResNet-$101$ & $79.3$ & - & - & - \\
				PSANet~\cite{ZhaoZLSLLJ18} & D-ResNet-$101$ & $80.1$ & - & - & - \\
				PADNet~\cite{OWS18} & D-ResNet-$101$ & $80.3$ & $58.8$ & $90.8$ & $78.5$\\
				CFNet~\cite{ZhangZWX} & D-ResNet-$101$ & $79.6$ & - & - & -\\
				Auto-DeepLab~\cite{liu2019auto} & - & $80.4$ & - & - & -\\
				DenseASPP~\cite{PSPNet} & WDenseNet-$161$ & $80.6$ & $59.1$ & $90.9$ & $78.1$ \\
				SVCNet~\cite{DingJSLW19} & ResNet-$101$ & $81.0$ & - & - & -\\
				ANNet~\cite{ZhuXBHB2019} & D-ResNet-$101$ & $81.3$ & - & - & -\\
				CCNet~\cite{HuangWHHWL2019} & D-ResNet-$101$ & $81.4$ & - & - & -\\
				DANet~\cite{DANet} & D-ResNet-$101$ & $81.5$ & - & - & - \\
				SFNet~\cite{li2020semantic} & D-ResNet-$101$ & $81.8$ & - & - & - \\
				\hline
				HRNetV$2$~\cite{hrnet} & HRNetV$2$-W$48$ & $81.6$ & $61.8$ & $\mathbf{92.1}$ & $\mathbf{82.2}$ \\
				U-HRNet &  U-HRNetV-W$48$ & $\mathbf{82.4}$ & $\mathbf{61.9}$ & $\mathbf{92.1}$ & $82.1$ \\
				\hline
				HRNetV$2$+OCR~\cite{YuanCW19} & HRNetV$2$-W$48$ & $82.5$ & $61.7$ & $\mathbf{92.1}$ & $81.6$ \\
				U-HRNet+OCR & U-HRNetV-W$48$ & $\mathbf{82.9}$ & $\mathbf{63.1}$ & $\mathbf{92.1}$ & $\mathbf{82.4}$ \\
				\hline
			\end{tabular}
		}
		\vspace{-2mm}
	\end{table}
	
	\begin{table}[t]
		\scriptsize
		\centering
		\caption{Semantic segmentation results on ADE20K \texttt{val} (multi-scale and flipping). (D- = Dilated-).}
		\label{tab:ade20kresults}
		\vspace{-.2cm}
		\resizebox{0.48\textwidth}{!}{
			\begin{tabular}{l|clc|ccc}
				\hline%
				& & backbone & & & mIoU & \\
				\hline
				\hline
				PSPNet~\cite{PSPNet} & & D-ResNet-$101$ & & & $43.29$ & \\
				PSANet~\cite{ZhaoZLSLLJ18} & & D-ResNet-$101$ & & & $43.77$ & \\
				EncNet~\cite{zhang2018context} & & D-ResNet-$101$ & & & $44.65$ & \\
				SFNet~\cite{li2020semantic} & & D-ResNet-$101$ & & & $44.67$ & \\
				CFNet~\cite{ZhangZWX} & & D-ResNet-$101$ & & & $44.89$ & \\
				CCNet~\cite{HuangWHHWL2019} & & D-ResNet-$101$ & & & $45.22$ & \\
				DANet~\cite{fu2019dual} & & D-ResNet-$101$ + multi-grid & & & $45.22$ & \\
				ANNet~\cite{ZhuXBHB2019} & & D-ResNet-$101$ & & & $45.24$ & \\
				APCNet~\cite{he2019adaptive} & & D-ResNet-$101$ & & & $45.38$ & \\
				ACNet\cite{fu2019adaptive} & & D-ResNet-$101$ + multi-grid & & & $45.90$ & \\
				\hline
				HRNet~\cite{hrnet} & & HRNetV$2$-W$48$ & & & $44.20$ & \\
				U-HRNet & & U-HRNet-W$48$ & & & $\mathbf{46.38}$ & \\
				\hline
				HRNet+OCR~\cite{YuanCW19} & & HRNetV$2$-W$48$ & & & $45.66$ & \\
				U-HRNet+OCR & & U-HRNet-W$48$ & & & $\mathbf{47.75}$ & \\
				\hline
			\end{tabular} 
		}
		\vspace{-.2cm}
	\end{table}
	
	\begin{table}[t]
		\centering
		\caption{Semantic segmentation results on LIP \texttt{val} (flipping). The overall performance of our approach is the best. (D- = Dilated-)}
		\label{tab:lipresults}
		\vspace{-.2cm}
		\resizebox{0.48\textwidth}{!}{
			\begin{tabular}{l|lc|ccc}
				\hline
				& backbone & extra. & pixel acc. & avg. acc. & mIoU \\
				\hline
				\hline
				Attention+SSL~\cite{GongLSL17} & VGG$16$ & Pose & $84.36$ & $54.94$ & $44.73$ \\
				DeepLabV$3$+~\cite{deeplabv3plus} & D-ResNet-$101$ & - & $84.09$ & $55.62$ & $44.80$ \\
				MMAN~\cite{LuoZZGYY18} & D-ResNet-$101$ & - & - & - & $46.81$ \\
				SS-NAN~\cite{ZhaoLNZCWFY17} & ResNet-$101$ & Pose & $87.59$ & $56.03$ & $47.92$ \\
				MuLA~\cite{NieFY18} & Hourglass & Pose & $\mathbf{88.50}$ & $60.50$ & $49.30$ \\
				JPPNet~\cite{XL18} & D-ResNet-$101$ & Pose & $86.39$ & $62.32$ & $51.37$ \\
				CE2P~\cite{TL18}  & D-ResNet-$101$ & Edge & $87.37$ & $63.20$ & $53.10$ \\
				\hline
				HRNetV$2$~\cite{hrnet} & HRNetV$2$-W$48$ & N & $88.21$ & $67.43$ & $55.90$ \\
				U-HRNet & U-HRNetV$2$-W$48$ & N & $88.34$ & $\mathbf{67.65}$ & $\mathbf{56.66}$ \\	
				\hline
				HRNetV$2$+OCR~\cite{YuanCW19} & HRNetV$2$-W$48$ & N & $88.24$ & $67.84$ & $56.48$ \\
				U-HRNet+OCR & U-HRNetV$2$-W$48$ & N & $88.34$ & $\mathbf{68.29}$ & $\mathbf{56.99}$ \\
				\hline
			\end{tabular}
		}
		\vspace{-2mm}
	\end{table}
	
	\begin{table}[t]
		\centering
		\caption{Semantic segmentation results on Synapse multi-organ CT dataset. The GFLOPs is calculated on the input size 224×224. The overall performance of our approach is the best.}
		\label{tab:synapseresults}
		\vspace{-.2cm}
		\resizebox{0.48\textwidth}{!}{
			\begin{tabular}{l|lc|cc}
				\hline
				& backbone & GFLOPs & average DSC $\uparrow$ & average HD $\downarrow$ \\
				\hline
				\hline
				V-Net~\cite{milletari2016fully} & V-Net & - & $68.81$ & $-$ \\
				DARR~\cite{fu2020domain} & V-Net & - & $69.77$ & $-$ \\
				U-Net~\cite{unet} & ResNet-50 & - & $74.68$ & $36.87$ \\
				AttnUNet~\cite{schlemper2019attention} & ResNet-50 & - & $75.57$ & $36.97$ \\
				TransUNet~\cite{chen2021transunet} & ResNet-50-ViT & $14.02$ & $77.48$ & $31.69$ \\
				\hline
				U-HRNet & U-HRNetV$2$-W$48$ & $17.01$ & $\mathbf{77.49}$ & $\mathbf{29.64}$ \\	
				\hline
			\end{tabular}
		}
		\vspace{-2mm}
	\end{table}
	
	\subsubsection{Comparison with state-of-the-art}
	
	Here we compare our proposed model with state-of-the-art methods. The proposed model is first pretrained on Imagenet and then fintuned on semantic semantic segmentation datasets. The configuration of training Imagenet is also kept same with ~\cite{hrnet}.
	
	\vspace{.1cm}
	\noindent\textbf{Results on the Cityscapes.}
	We report the results of U-HRNet and other state-of-the-art methods on Cityscapes \textit{val} set in terms of GFLOPs and mIoU in Table~\ref{tab:cityscapevalresults}: (i) U-HRNet-W18-small which has similar GFLOPs as HRNetV$2$-W18-small, achieve $78.5$ mIoU, $2.2$ points gain over HRNetV$2$-W18-small-v2, and outperforms other small models by large margins with less calculation. (ii) U-HRNet-W48 which has similar GFLOPs as HRNetV$2$-W48, achieve $81.9$ mIoU, $0.8$ points gain over HRNetV$2$-W48, and also outperforms other main stream large models by large margins with much lower computation cost.
	
	Table~\ref{tab:cityscaperesults} provides the comparison of U-HRNet with state-of-the-arts methods on Cityscapes \textit{test} set. There are two situations, learning on \textit{train} set and learning on \textit{train}+\textit{val} set. In both situations, U-HRNet-W$48$ preforms better than HRNetV$2$-W$48$ and other state-of-the-arts methods.
	
	It is worth mentioning that, U-HRNet-W$48$ achieves comparable mIoU with HRNetV$2$-W$48$+OCR on both \textit{val} and \textit{test} sets by using only $57.9\%(698.6/1206.3)$ GFLOPs. Even more, by adding OCR module, it can further achieve $82.9$ mIoU on \textit{test} set which sets a new state-of-the-art.
	
	\noindent\textbf{Results on the ADE20K.}
	Table~\ref{tab:ade20kresults} illustrates the comparison of our proposed method with state-of-the-art methods on ADE20K \textit{val} set. U-HRNet-W$48$ outperforms HRNetV$2$-W$48$ by a large margin of $2.18$ points, and performs better than other state-of-the-arts methods as well. Further more, U-HRNet-W$48$+OCR achieves $47.75$ mIoU, which pushes the state-of-the-art forward significantly.
	
	\noindent\textbf{Results on the LIP.}
	Table~\ref{tab:lipresults} shows the comparison of U-HRNet with state-of-the-art methods on LIP \textit{val} set. U-HRNet-W$48$ get $0.76$ points gain over HRNetV$2$-W$48$ on mIoU with a similar computation cost, and performs better than other methods as well, without using extra information, such as pose and edges. While, U-HRNet-W$48$+OCR achieves $56.99$ mIoU which is also a new state-of-the-art.
	
	\noindent\textbf{Results on the Synapse multi-organ CT.}
	As shown in Table~\ref{tab:synapseresults}, U-HRNet-W$48$ outperforms U-Net series networks significantly. Especially, comparing to the recent TransUNet which is transformer based, U-HRNet-W$48$ gets $2.05$ mm improvement only with a few GFLOPs increasement. In addition, U-HRNet is fully convolutional without outer product operation between tensors, which is more friendly for computation than transformer based network.

	\section{Depth Estimation}
	
	Depth estimation is a problem of predicting depth value for each pixel in a image. It is a typical dense regression task in scene understanding. Here, we carry out a certain amount of experiments on a widely used dataset, NYUDv2, and exhibit competitive results.
	
	\subsection{Dataset}
	
	\noindent\textbf{NYUD-V2.} The NYU Depth V2 (NYUD-V2) dataset  contains 120K RGB-depth pairs with a $480\times640$ size, which is acquired by Microsoft Kinect from 464 different indoor scenes. 
	Apart from the whole dataset, there are officially annotated $1449$ indoor images (NYUD-Small), in which $795$ images are split for training.
	Following previous works~\cite{lee2019big,wei2019enforcing}, we using other $654$ images as test set throughout all of the experiment of depth estimation and eigen crop is conducted. 
	In order to verify the scalibility of our method, we use a large dataset, named as NYUD-Large in this paper, for training in addition. This dataset contains $24231$ RGB-depth pairs released by ~\cite{lee2019big}.
	
	\begin{table}[!t]
		\centering
		\caption{Depth estimation results on NYUD-Small and NYUD-Large dataset. With our U-HRNet backbone, the performance is improved over all evaluation metrics.}
		\vspace{-.3cm}
		\resizebox{0.48\textwidth}{!}{
			\begin{tabular}{l|ccc|ccc}
				\hline
				\multirow{2}{*}{Method} & abs-rel & $log_{10}$ & rms & $\delta_{1}$ & $\delta_{2}$ & $\delta_{3}$ \\
				& \multicolumn{3}{c|}{(Lower is better)} & \multicolumn{3}{c}{(Higher is better)} \\ 
				\hline
				\hline
				\multicolumn{7}{c}{NYUD-Small} \\
				\hline
				HRNetV$2$-W18-small~\cite{hrnet} & $0.186$  & $0.076$ & $0.591$ & $0.719$ & $0.933$ & $0.984$ \\
				U-HRNet-W18-small & $\mathbf{0.172}$ & $\mathbf{0.068}$ & $\mathbf{0.534}$ & $\mathbf{0.763}$ & $\mathbf{0.950}$ & $\mathbf{0.988}$ \\
				\hline
				HRNetV$2$-W48~\cite{hrnet} & $0.156$ & $0.064$ & $0.510$ & $0.789$ & $0.957$ & $0.991$ \\
				U-HRNet-W48 & $\mathbf{0.150}$ & $\mathbf{0.061}$ & $\mathbf{0.484}$ & $\mathbf{0.811}$ & $\mathbf{0.960}$ & $0.991$ \\
				\hline
				\hline
				\multicolumn{7}{c}{NYUD-Large} \\
				\hline
				HRNetV$2$-W18-small~\cite{hrnet} & $0.147$ & $0.062$ & $0.500$ & $0.798$ & $0.958$ & $0.990$ \\
				U-HRNet-W18-small & $\mathbf{0.127}$ & $\mathbf{0.054}$ & $\mathbf{0.456}$ & $\mathbf{0.849}$ & $\mathbf{0.968}$ & $0.990$ \\
				\hline
				HRNetV$2$-W48~\cite{hrnet} & $0.123$ & $0.053$ & $0.448$ & $0.846$ & $0.968$ & $0.991$ \\
				U-HRNet-W48 & $\mathbf{0.117}$  & $\mathbf{0.051}$ & $\mathbf{0.440}$ & $\mathbf{0.863}$ & $\mathbf{0.970}$ & $0.991$ \\
				\hline
			\end{tabular}
		}
		\label{table:nyud-cmp}
		\vspace{-.5cm}
	\end{table}
	
	\subsection{Implementation Details}
	
	\noindent\textbf{Network structures.}
	We apply the same network structure as used in semantic segmentation, with only the number of output channel of the last convolution adapting to depth estimation, as same as the implementation in~\cite{vandenhende2020mti}.
	
	\noindent\textbf{Training details.}
	The same data augmentation strategy as described in ~\cite{vandenhende2020mti} is used, the
	RGB-depth pairs are randomly scaled with the selected ratio in {1, 1.2, 1.5} and randomly horizontally flipped. For training configuration, Adam optimizer with initial learning rate 1e-4 and weight decay 1e-4 is applied, while polynomial schedule is employed for learning rate decay. For NYUD-Small, the bath size and total number of epochs are set to $6$ and $80$ respectively, while $16$ and $50$ for NYUD-Large. Imagenet pretraining is carried out for all of the experiments.
	
	\subsection{Evaluation Metrics}
	
	Following previous methods~\cite{alhashim2018high,wei2019enforcing}, we use six common metrics to evaluate the performance of monocular depth estimation quantitatively: mean absolute relative error (abs-rel), mean $\log_{10}$ error ($\log_{10}$), root mean squared error (rms) , and the accuracy under threshold ($\delta_{i} < 1.25^{i}, i=1, 2, 3$). 
	
	\subsection{Experimental Results}
	
	\begin{table}[!t]
		\caption{Compare with state-of-the-arts on NYUD. Our approach performances better than most of works except for Wei \etal.~\cite{wei2019enforcing} that developed a strong 3D-based method.}
		\vspace{-.3cm}
		\resizebox{0.47\textwidth}{!}{
			\begin{tabular}{l|ccc|ccc}
				\hline
				\multirow{2}{*}{Method} & abs-rel & $log_{10}$ & rms & $\delta_{1}$ & $\delta_{2}$ & $\delta_{3}$ \\
				& \multicolumn{3}{c|}{(Lower is better)}         & \multicolumn{3}{c}{(Higher is better)} \\ 
				\hline
				\hline
				Eigen \etal.~\cite{eigen2015predicting}  & $0.158$  & -     & $0.641$ & $0.769$  & $0.950$  & $0.988$ \\
				Chakrabarti \etal.~\cite{chakrabarti2016depth}      & $0.149$  & - & $0.620$  & $0.806$  & $0.958$  & $0.987$ \\
				Li \etal.~\cite{li2017two}   & $0.143$   & $0.063$  & $0.635$  & $0.788$  & $0.958$ & $0.991$    \\
				Laina \etal.~\cite{laina2016deeper}   & $0.127$  & $0.055$    & $0.573$  & $0.811$   & $0.953$  & $0.988$      \\
				Fu \etal.~\cite{fu2018deep}   & $0.115$  & $0.051$  & $0.509$   & $0.828$  & $0.965$  & $0.992$  \\   
				Hu \etal.~\cite{hu2019revisiting} & $0.115$  & $0.050$    & $0.530$  & $0.866$   & $0.975$  & $0.993$      \\
				Zhang \etal.~\cite{zhang2019pattern} & $0.121$ & - & $0.497$ & $0.846$ & $0.968$ & $0.994$ \\
				Wang \etal.~\cite{wang2020sdc} & $0.128$ & - & $0.497$ & $0.845$ & $0.966$ & $0.990$ \\
				Chen \etal.~\cite{chen2019attention} & $0.138$ & - & $0.496$ & $0.826$ & $0.964$ & $0.990$ \\
				Alhashim \etal.~\cite{alhashim2018high} & $0.123$  & $0.053$    & $0.465$  & $0.846$   & $0.974$  & $0.994$      \\
				\hline
				Wei \etal.~\cite{wei2019enforcing} & $\mathbf{0.108}$    & $\mathbf{0.048}$   & $\mathbf{0.416}$   & $\mathbf{0.875}$      & $\mathbf{0.976}$    & $0.994$ \\
				\hline
				U-HRNet-W18-small (Ours)    & $0.127$    & $0.054$   & $0.456$   & $0.849$           & $0.968$    & $0.990$   \\ 
				U-HRNet-W48 (Ours)    & $0.117$    & $0.051$   & $0.440$   & $0.863$           & $0.970$    & $0.991$   \\ 
				\hline
			\end{tabular}
		}
		\label{table:nyud-sota}
		\vspace{-.5cm}
	\end{table}
	
	As shown in Table~\ref{table:nyud-cmp}, both on NYUD-Small and NYUD-Large, U-HRNet outperforms HRNetV$2$ by a remarkable margin, in particular of the small model, which outperforms the baseline HRNetV$2$-W18-small at rmse by 0.057 and 0.044 on NYUD-small and NYUD-Large respectively. Meanwhile, our method is also competitive with the state-of-the-art methods on NYUDv2. As depicted in Table~\ref{table:nyud-sota}, our U-HRNet-W48 achieve a rmse of 0.440 which is better than most of the previous methods. And more impressively, U-HRNet-W18-small also get a very competitive rms of 0.456 without any other additional tricks or modules for improvement. These all indicate that our model can also work well on dense regression task.
	
	\section{Conclusion}
	In this paper, we present a U-Shape high resolution network for dense prediction tasks. It has two fundamental differences from the existing high resolution network: (i) U-HRNet adds more stages after the feature map with strongest semantic representation which enables this representation can be fully utilized for further inference. (ii) U-HRNet relaxes the constraint that all resolutions need to be calculated parallel for a newly added stage, which makes the network can assign more calculations on low-resolution stages and get stronger semantic representation. U-HRNet has been verified to be more effective on several datasets of semantic segmentation and depth estimation than the existing high resolution network with a similar computation cost, and we will explore on more other dense prediction tasks such as super-resolution, inpainting, image enhancement and so on.
	
	{\small
		\bibliographystyle{ieee_fullname}
		\bibliography{refs}
	}
	
\end{document}